\documentclass[runningheads]{llncs}

\usepackage{amsmath}
\usepackage{amssymb}                                
\usepackage{bm}                                     
\usepackage{booktabs}                               
\usepackage{subcaption}                             
\usepackage{comment}                                
\usepackage[nolist]{acronym}                        
\usepackage{graphicx}                               
\usepackage[utf8]{inputenc}                         
\usepackage{mathtools}                              
\usepackage{siunitx}                                
\usepackage{xcolor}           

\usepackage{hyperref}
\usepackage[capitalise]{cleveref}                   

\newcommand{\comrem}[1]{}

\newcommand{\myhref}[1]{\href{#1}{\texttt{#1}}}

\newcommand{\x}{\mathbf{x}}

\newcommand{\z}{\mathbf{z}}

\newcommand{\sss}{\mathbf{s}}

\newcommand{\R}{\mathbb{R}}

\newcommand{\F}{\mathcal{F}}
\newcommand{\C}{\mathcal{C}}
\newcommand{\RR}{\mathcal{R}}

\newcommand{\Ypred}{\hat{Y}}
\newcommand{\Yb}{\mathbf{Y}}

\newcommand{\thetab}{\bm{\theta}}
\newcommand{\pars}[1]{\thetab_{#1}}
\newcommand{\FC}[1]{\F_{\pars{#1}}}
\newcommand{\CNN}[1]{\C_{\pars{#1}}}
\newcommand{\RNN}[1]{\RR_{\pars{#1}}}
\newcommand{\Cx}{\CNN{\x}}

\newcommand{\Fzx}{\FC{\z,\x}}

\newcommand{\Fzs}{\FC{\z,\sss}}
\newcommand{\Rs}{\RNN{\sss}}
\newcommand{\st}[1]{_{\text{#1}}}

\makeatletter
\newcommand*{\org@overidelabel}{}
\let\org@overridelabel\@verridelabel
\@ifpackagelater{acronym}{2015/03/21}{
  \renewcommand*{\@verridelabel}[1]{%
    \@bsphack
    \protected@write\@auxout{}{\string\AC@undonewlabel{#1@cref}}%
    \org@overridelabel{#1}%
    \@esphack
  }%
}{
  \renewcommand*{\@verridelabel}[1]{%
    \@bsphack
    \protected@write\@auxout{}{\string\undonewlabel{#1@cref}}%
    \org@overridelabel{#1}%
    \@esphack
  }%
}
\makeatother

\graphicspath{{./figures/}}

\title{\LARGE \bf
    Transfer Learning of Deep Spatiotemporal Networks to Model Arbitrarily Long Videos of Seizures
}
\titlerunning{Transfer Learning to Model Arbitrarily Long Videos of Seizures}

\author{%
    Fernando Pérez-García\inst{1,2,3} \and  
    Catherine Scott\inst{4} \and                                         
    Rachel Sparks\inst{3} \and              
    Beate Diehl\inst{4} \and                
    Sébastien Ourselin\inst{3} \\           
}

\authorrunning{F. Pérez-García et al.}

\institute{
    Department of Medical Physics and Biomedical Engineering, University College London, UK \and
    Wellcome / EPSRC Centre for Interventional and Surgical Sciences (WEISS), University College London, UK \and
    School of Biomedical Engineering \& Imaging Sciences (BMEIS), King's College London, UK \and
    UCL Queen Square Institute of Neurology, Department of Clinical and Experimental Epilepsy, London, UK \and
    National Hospital for Neurology and Neurosurgery, Department of Clinical Neurophysiology, London, UK \\
    \email{fernando.perezgarcia.17@ucl.ac.uk}
}

\begin{acronym}
    \acro{cnn}[CNN]{convolutional neural network}
    \acro{eeg}[EEG]{electroencephalography}
    \acro{emu}[EMU]{epilepsy monitoring unit}
    \acro{ez}[EZ]{epileptogenic zone}
    \acro{tcs}[TCS]{focal to bilateral tonic-clonic seizure}
    \acro{fos}[FOS]{focal onset seizure}
    \acro{fov}[FOV]{field of view}
    \acro{fps}[FPS]{frames per second}
    \acro{gesture}[GESTURES]{Generalized Epileptic Seizure classification from video-Telemetry Using REcurrent convolutional neural networkS}
    \acro{gtcs}[GTCS]{generalized tonic-clonic seizure}
    \acro{har}[HAR]{human action recognition}
    \acro{hevc}[HEVC]{High Efficiency Video Coding}
    \acro{lstm}[LSTM]{long short-term memory}
    \acro{blstm}[BLSTM]{bidirectional \ac{lstm}}
    \acro{blstmac}[BLSTM]{bidirectional long short-term memory}
    \acro{pges}[PGES]{postictal generalized \ac{eeg} suppression}
    \acro{rnn}[RNN]{recurrent neural network}
    \acro{roi}[ROI]{region of interest}
    \acrodefplural{roi}[ROI]{regions of interest}
    \acro{sfcnn}[SFCNN]{single-frame \ac{cnn}}
    \acro{stcnn}[STCNN]{spatiotemporal \ac{cnn}}
    \acro{sudep}[SUDEP]{sudden unexpected death in epilepsy}
    \acro{tsn}[TSN]{temporal segment network}

    \acro{fle}[FLE]{frontal lobe epilepsy}
    \acro{tle}[TLE]{temporal lobe epilepsy}
    \acro{mtle}[MTLE]{mesial temporal lobe epilepsy}
    \acro{etle}[ETLE]{extratemporal lobe epilepsy}
\end{acronym}

\begin{document}
    \maketitle
    \begin{abstract}

    Detailed analysis of seizure semiology, the symptoms and signs which occur during a seizure, is critical for management of epilepsy patients.
    Inter-rater reliability using qualitative visual analysis is often poor for semiological features.
    Therefore, automatic and quantitative analysis of video-recorded seizures is needed for objective assessment.
    We present \acs{gesture}, a novel architecture combining \acp{cnn} and \acp{rnn} to learn deep representations of arbitrarily long videos of epileptic seizures.
    We use a \ac{stcnn} pre-trained on large \ac{har} datasets to extract features from short snippets ($\approx \SI{0.5}{\second}$) sampled from seizure videos.
    We then train an \ac{rnn} to learn seizure-level representations from the sequence of features.
    We curated a dataset of seizure videos from 68 patients and evaluated \acs{gesture} on its ability to classify seizures into \acp{fos} ($N = 106$) vs. \acp{tcs} ($N = 77$), obtaining an accuracy of 98.9\% using \ac{blstmac} units.
    We demonstrate that an \ac{stcnn} trained on a \ac{har} dataset can be used in combination with an \ac{rnn} to accurately represent arbitrarily long videos of seizures.
    \acs{gesture} can provide accurate seizure classification by modeling sequences of semiologies.
    The code, models and features dataset are available at \myhref{https://github.com/fepegar/gestures-miccai-2021}.

\end{abstract}

\acresetall

    \section{Introduction}


Epilepsy is a neurological condition characterized by abnormal brain activity that gives rise to seizures, affecting about 50 million people worldwide~\cite{fiest_prevalence_2017}.
Seizure semiology, ``the historical elicitation or observation of certain symptoms and signs'' during seizures, provides context to infer epilepsy type~\cite{fisher_operational_2017}.
\Acp{fos} start in a region of one hemisphere.
If they spread to both hemispheres, they are said to \emph{generalize}, becoming \acp{tcs}~\cite{fisher_operational_2017}.
In \acp{tcs}, the patient first presents semiologies associated with a \ac{fos}, such as head turning
or mouth and hand automatisms.  
This is followed by a series of phases, in which muscles stiffen (tonic phase) and limbs jerk rapidly and rhythmically (clonic phase).
\Acp{tcs} put patients at risk of injury and, if the seizure does not self-terminate rapidly, can result in a medical emergency.
\acs{sudep} is the sudden and unexpected death of a patient with epilepsy, without evidence of typical causes of death. \acused{sudep}
Risk of \ac{sudep} depends on epilepsy and seizure characteristics as well as living conditions.
\Acp{tcs} in particular increase \ac{sudep} risk substantially~\cite{nashef_unifying_2012}. 
In a small number of \ac{sudep} cases occurring in \acp{emu}, death was preceded by a \ac{tcs} followed by cardiorespiratory dysfunction minutes after seizure offset~\cite{ryvlin_incidence_2013}.
Identifying semiologies related to increased risk of \ac{sudep} to appropriately target treatment is an open research question.
One limitation determining \ac{sudep} risk factors is that inter-rater reliability based on qualitative visual analysis is poor for most semiological features (e.g., limb movement, head pose or eye gaze), especially between observers from different epilepsy centers~\cite{tufenkjian_seizure_2012}.
Therefore, automatic and quantitative analysis of video-recorded seizures is needed to standardize assessment of seizure semiology across multicenter studies~\cite{ahmedtaristizabal_automated_2017}.


Early quantitative analysis studies of epileptic seizures evaluated patient motion by attaching infrared reflective markers to key points on the body or using cameras with color and depth streams~\cite{li_z_movement_2002,cunha_movement_2003,odwyer_lateralizing_2007,cunha_neurokinect_2016}.
These methods are not robust to occlusion by bed linens or clinical staff, differences in illumination and pose, or poor video quality caused by compression artifacts or details out of focus.

Neural networks can overcome these challenges by automatically learning features from the training data that are more robust to variations in the data distribution.
Most related works using neural networks focus on classifying the \emph{epilepsy type} by predicting the location of the \ac{ez}, e.g., ``\acl{tle}'' vs. ``\acl{etle}'', from short ($\le \SI{2}{s}$) snippets extracted from videos of one or more seizures~\cite{ahmedt-aristizabal_deep_2018,ahmedt-aristizabal_hierarchical_2018,ahmedt-aristizabal_deep_2018-1,maia_epileptic_2019,karacsony_deep_2020}.
Typically, this is done as follows.
First, the bed is detected in the first frame and the entire video is cropped so the \ac{fov} is centered on the bed.
During training, a \ac{cnn} is used to extract features for each frame in a sampled snippet.
Then, a \ac{rnn} aggregates the features into a \emph{snippet-level} representation and a fully-connected layer predicts the epilepsy type.
Finally, a \emph{subject-level} prediction is obtained by averaging all snippet-level predictions.
This approach has several disadvantages.
First, it is not robust to incorrect bed detection or changes in the \ac{fov} due to zooming or panning.
Second, the order of semiologies is ignored, as the epilepsy type is predicted from short snippets independently of their occurrence during a seizure.
Moreover, patients with the same epilepsy type may present different seizure types.
Finally, training neural networks from small datasets, as is often the case in clinical settings, leads to limited results.

The goal of this work is to compute \emph{seizure-level} representations of arbitrarily long videos when a small dataset is available, which is typically the case in \acp{emu}.

To overcome the challenge of training with small datasets, transfer learning from \acp{stcnn} trained for \ac{har} can be used~\cite{karacsony_deep_2020}.
Although seizures are, strictly speaking, not actions, \ac{har} models are expected to encode strong representations of human motion that may be relevant for seizure characterization.
These methods are typically designed to classify human actions by aggregating predictions for snippets sampled from short clips ($\approx \SI{10}{\second}$).
Epileptic seizures, however, can last from seconds to tens of minutes~\cite{jenssen_how_2006}.
A common aggregation method is to average predictions from randomly sampled snippets~\cite{carreira_quo_2017,ghadiyaram_large-scale_2019,simonyan_two-stream_2014}.
Averaging predictions typically works because most video datasets considered are trimmed, i.e., the same action occurs along most of the video duration.
In our dataset, due to the nature of \acp{tcs}, more than half the frames are labeled as non-generalizing in 49/79 (62\%) of the \ac{tcs} videos.
Therefore, simply averaging snippet-level predictions would result in a large number of seizures being misclassified as \acp{fos}.
\Acp{tsn}~\cite{wang_temporal_2019} split videos of any duration into $n$ non-overlapping segments and a consensus function aggregates features extracted from each segment.
Therefore, we propose the use of \acp{tsn} to capture semiological features across the entirety of the seizure.
We use an \ac{rnn} as a consensus function to model the sequence of feature vectors extracted from the segments.

We present a novel neural network architecture combining \acp{tsn} and \acp{rnn}, which we denote \ac{gesture}, that provides full representations of arbitrarily long seizure videos.
These representations could be used for tasks such as classification of seizure types, seizure description using natural language, or triage.
To model the relevant patient motion during seizure without the need for object detection, we use a \ac{stcnn} trained on large-scale \ac{har} datasets (over 65 million videos from Instagram and 250,000 from YouTube)~\cite{ghadiyaram_large-scale_2019} to extract features from short snippets.
Then, an \ac{rnn} is used to learn a representation for the full duration of the seizure.

We chose as a proof of concept to distinguish between \acp{fos} and \acp{tcs}, because the key distinction, if the discharge spreads across hemispheres, is only observed later in the seizure.
This task demonstrates that we can train a model to take into account features across the entirety of the seizure.
The main challenge, apart from the typical challenges in video-telemetry data described above, is distinguishing between \acp{tcs} and hyperkinetic \acp{fos}, which are characterized by intense motor activity involving the extremities and trunk.

\section{Materials and methods}

\subsection{Video acquisition}

Patients were recorded using two full high-definition ($1920 \times 1080$ pixels, 30 \ac{fps}) cameras installed in the \ac{emu} as part of standard clinical practice.
Infrared is used for acquisition in scenes with low light intensity, such as during nighttime.
The acquisition software (Micromed, Treviso, Italy) automatically resizes one of the video streams ($800 \times 450$), superimposes it onto the top-left corner of the other stream and stores the montage using MPEG-2.
See the supplementary materials for six examples of videos in our dataset.

\subsection{Dataset description and ground-truth definitions}
\label{sec:dataset}

A neurophysiologist (A.A.) annotated for each seizure the following times: clinical seizure onset $t_0$, onset of the clonic phase $t_G$ (\acp{tcs} only) and clinical seizure offset $t_1$.
The annotations were confirmed using \ac{eeg}.

We curated a dataset comprising 141 \acp{fos} and 77 \acp{tcs} videos from 68 epileptic patients undergoing presurgical evaluation at the National Hospital for Neurology and Neurosurgery, London, United Kingdom.
To reduce the seizure class imbalance, we discarded seizures where $t_1 - t_0 < \SI{15}{\second}$, as this threshold is well under the shortest reported time for \acp{tcs}~\cite{jenssen_how_2006}.
After discarding short videos,
there were 106 \acp{fos}.
The `median (min, max)' number of seizures per patient is 2 (1, 16).
The duration of \ac{fos} and \ac{tcs} is 53 (16, 701) s and 93 (51, 1098) s, respectively.
The total duration of the dataset is 298 minutes, 20\% of which correspond to \ac{tcs} phase (i.e., the time interval $[t_G, t_1]$).
Two patients had only \ac{fos}, 32 patients had only \ac{tcs}, and 34 had seizures of both types.
The `mean (standard deviation)' of the percentage of the seizure duration before the appearance of generalizing semiology, i.e., $r = (t_G - t_0) / (t_1 - t_0)$, is 0.56 (0.18), indicating that patients typically present generalizing semiological features in the second half of the seizure.



Let a seizure video be a sequence of $K$ frames starting at $t_0$.
%
Let the time of frame $k \in \{ 0, \dots, K - 1 \}$ be ${t_k = t_0 + \frac{k}{f}}$, where $f$ is the video frame rate.
%
We use 0 and 1 to represent \ac{fos} and \ac{tcs} labels, respectively.
The ground-truth label $y_k \in \{0, 1\}$ for frame $k$ is defined as
$y_k \coloneqq 0$ if $t_k < t_G$ and 1 otherwise,
where $t_G \rightarrow \infty$ for \acp{fos}.

Let $\x \in \R ^ {3 \times l \times h \times w}$ be a stack of frames or \textit{snippet}, where
$3$ denotes the RGB channels,
$l$ is the number of frames,
and $h$ and $w$ are the number of rows and columns in a frame, respectively.
The label for a snippet starting at frame $k$ is
\begin{equation}
    Y_k \coloneqq
    \left\{
        \begin{array}{ll}
            0 & \mbox{if } \frac{t_k + t_{k + l}}{2} < t_G \\
            1 & \mbox{otherwise}
        \end{array}
    \right.
\end{equation}

\subsection{Snippet-level classification}
\label{sec:snippet-level}




The probability $\hat{Y}_k$ that a patient presents generalizing features within snippet $\x_k$ starting at frame $k$ is computed as
\begin{equation}
    \Ypred_k = \Pr(Y_k = 1 \mid \x_k) = \Fzx( \Cx(\x_k) ) = \Fzx( \z_k)
\end{equation}
where
$\Cx$ is an \ac{stcnn} parameterized by $\pars{\x}$ that extracts features,
$\z_k \in \R ^ m$ is a vector of $m$ features representing $\x_k$ in a latent space,
and
$\Fzx$ is a fully-connected layer parameterized by $\pars{\z,\x}$ followed by a sigmoid function that maps logits to probabilities.
In this work, we do not update $\pars{\x}$ during training.

\subsection{Seizure-level classification}
\label{sec:meth_seizure}

\subsubsection{Temporal segment network}
\begin{figure}[t!]
    \centering
    \begin{subfigure}[b]{0.49\textwidth}
        \centering
        \includegraphics[width=\textwidth]{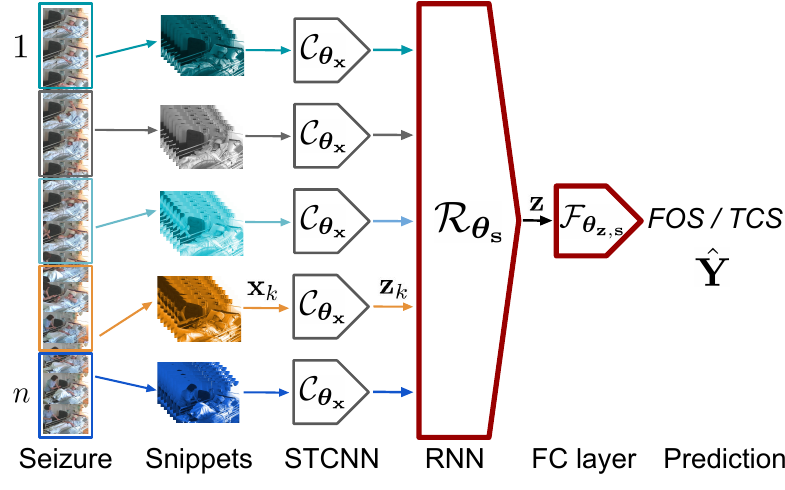}
        \caption{\ac{gesture} architecture}
        \label{fig:diagram}
    \end{subfigure}
    \hfill
    \begin{subfigure}[b]{0.49\textwidth}
        \centering
        \includegraphics[width=\textwidth]{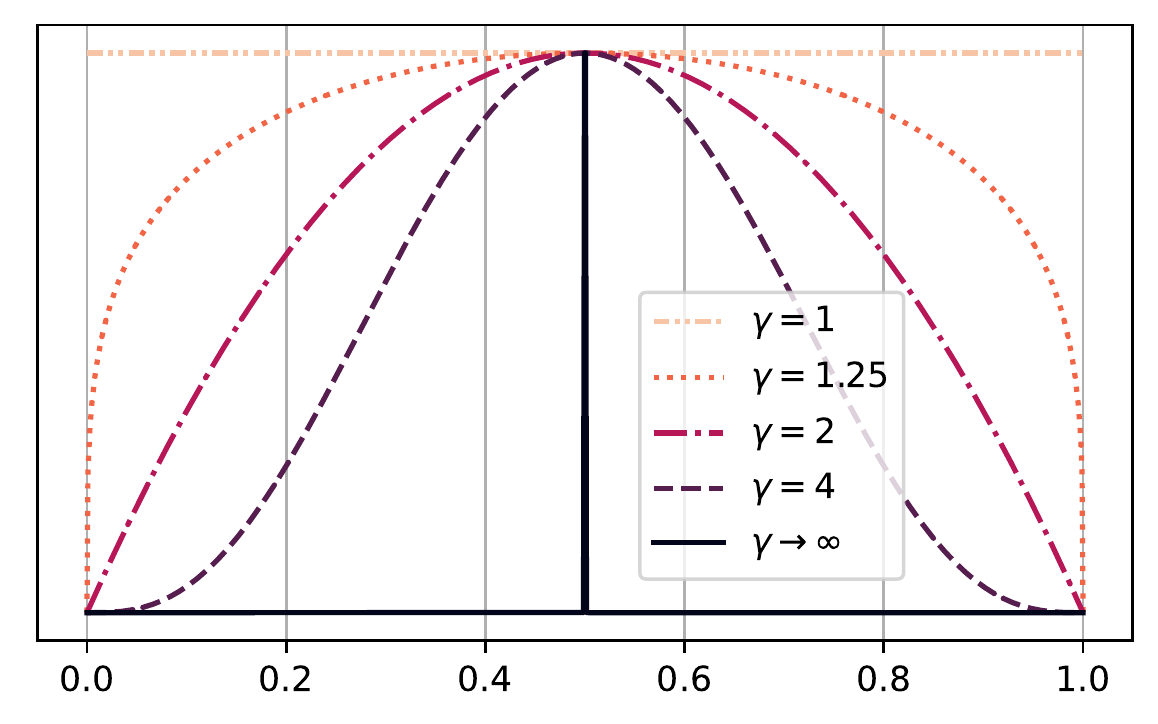}
        \caption{Sampling distributions}
        \label{fig:betas}
    \end{subfigure}
    \caption{%
        Left: The \ac{gesture} architecture.
        We train only the models with thick red borders.
        Right: Probability distributions used to sample snippets from video segments.
    }
    \label{fig:diagram_betas}
\end{figure}

Let $V = \{ \x_k \}_{k=1}^{K-l}$ be the set of all possible snippets sampled from a seizure video.
We define a sampling function $f : (V, n, \gamma) \mapsto S$ that extracts a sequence $S$ of $n$ snippets by splitting $V$ into $n$ non-overlapping segments and randomly sampling one snippet per segment.
There are two design choices: the number of segments $n$ and the probability distribution used for sampling within a segment.
If a uniform distribution is used,
information from two adjacent segments might be redundant.
Using the middle snippet of a segment minimizes redundancy, but reduces the proportion of data leveraged during training.
We propose using a symmetric beta distribution ($\text{Beta}(\gamma, \gamma)$) to model the sampling function,
where $\gamma$ controls the dispersion of the probability distribution (\cref{fig:betas}).
The set of latent snippet representations is $Z = \{ \Cx (\x_i) \}_{i = 1}^{n}$.

\subsubsection{Recurrent neural network}

To perform a seizure-level prediction $\hat{\Yb}$, $Z$ is aggregated as follows:
\begin{equation}
    \hat{\Yb}
    = \Pr(\Yb = 1 \mid S)
    = \Fzs( \Rs ( Z ) )
    = \Fzs( \z )
\end{equation}
where
$\Rs$ is an \ac{rnn} parameterized by $\pars{\sss}$,
$\Fzs$ is a fully-connected layer parameterized by $\pars{\z,\sss}$ which uses a softmax function to output probabilities,
and $\z$ is a feature-vector representation of the entire seizure video, corresponding to the last hidden state of $\Rs$.

\section{Experiments and results}

All videos were preprocessed by
separating the two streams into different files (replacing the small embedded view with black pixels),
resampling to 15 \ac{fps} and $320 \times 180$ pixels,
and reencoding using \ac{hevc}.
To avoid geometric distortions while maximizing the \ac{fov} and resolution, videos were cropped horizontally by removing 5\% of the columns from each side, and padded vertically so frames were square.
Snippets were resized to $224 \times 224$ or $112 \times 112$, as imposed by each architecture.
For realism, six video streams in which the patient was completely outside of the \ac{fov} were discarded for training but used for evaluation.

Experiments were implemented in PyTorch 1.7.0.
We used a stratified 10-fold cross-validation, generated to ensure the total duration of the videos and ratio of \acp{fos} to \acp{tcs} were similar across folds.
Both views from the same video were assigned to the same fold, but videos from the same patient were not.
This is because individual patients can present with both \acp{fos} or \acp{tcs}, so data leakage at the patient level is not a concern.
We minimized the weighted binary cross-entropy loss to overcome dataset imbalance, using the AdamW optimizer~\cite{loshchilov_decoupled_2019}.
The code is available at \myhref{https://github.com/fepegar/gestures-miccai-2021}.

For each fold, evaluation is performed using the model from the epoch with the lowest validation loss.
At inference time, the network predicts probabilities for both video streams of a seizure, and these predictions are averaged.
The final binary prediction is the consensus probability thresholded at 0.5.
We analyzed differences in model performance using a one-tailed Mann-Whitney $U$ test (as metrics were not normally distributed) with a significance threshold of $\alpha = 0.05$, and Bonferroni correction for each set of $e$ experiments: $\alpha\st{Bonf} = \frac{\alpha}{e (e - 1)}$.

\subsection{Evaluation of feature extractors for snippets}
\label{sec:exp_feat}

Despite recent advances in \acp{stcnn} for \ac{har}, these architectures do not always outperform \acp{sfcnn} pre-trained on large generic datasets~\cite{hutchinson_accuracy_2020}.
We assessed the ability of different feature extractors to model semiologies by training a classifier for snippet-level classification (\cref{sec:snippet-level}).

\begin{table}[t!]
    \setlength{\tabcolsep}{3pt}
    \centering
    \caption{%
        Performance of the feature extractors.
        The number of parameters is shown in millions.
        AUC is the area under the precision-recall curve.
        Accuracy is computed for \acp{tcs} and \acp{fos}, while $F_1$-score and AUC only for \acp{tcs}, represented by an asterisk (*).
        Metrics are expressed as `median (interquartile range)'.
    }
    \label{tab:models}
    \begin{tabular}{l*5c}
        \toprule
        \textbf{Model (frames)} & \textbf{Parameters} & \textbf{Features} & \textbf{Accuracy} & $\bm{F_1}$\textbf{-score}* & \textbf{AUC}* \\
        \midrule
        Wide R2D-50-2 (1)       &              66.8 M &              2048 &       80.3 (33.2) &                67.4 (30.2) &   75.7 (38.4) \\
        R2D-34 (1)              &              21.2 M &               512 &       89.7 (27.7) &                73.9 (23.6) &   84.3 (28.7) \\
        R(2+1)D-34 (8)          &              63.5 M &               512 &       93.9 (18.3) &                81.6 (16.9) &   93.7 (13.4) \\
        R(2+1)D-34 (32)         &              63.5 M &               512 &       96.9 (12.9) &                84.7 (13.4) &   94.7 (11.9) \\
        \bottomrule
    \end{tabular}
\end{table}

We used two pre-trained versions of the \ac{stcnn} R(2+1)D-34~\cite{ghadiyaram_large-scale_2019} that take as inputs 8 frames ($\approx \SI{0.5}{s}$) or 32 frames ($\approx \SI{2.1}{s}$).
Models were trained using weakly supervised learning on over 65 million Instagram videos and fully supervised learning on over 250,000 YouTube videos of human actions.
We selected two pre-trained \acp{sfcnn} with 34 (R2D-34) and 50 (Wide R2D-50-2) layers, trained on ImageNet~\cite{zagoruyko_wide_2017}.
The \acp{sfcnn} were chosen so the numbers of layers (34) and parameters ($\approx 65$ million) were similar to the \acp{stcnn}.

To ensure that all features datasets have the same number of training instances, we divided each video into segments of 32 frames.
Then, we use the models to extract features from snippets of the required length (8, 32, or 1) such that all snippets are centered in the segments.
The datasets of extracted feature vectors are publicly available~\cite{perez-garcia_data_2021}.
We trained a fully-connected layer for 400 epochs on each feature set, treating views from the same video independently.
We used an initial learning rate $10 ^ {-3}$ and mini-batches of 1024 feature vectors.
We minimized a weighted binary cross-entropy loss,
where the weight for \acp{tcs} was computed as the ratio of \ac{fos} frames to \ac{tcs} frames.

For evaluation, a sliding window was used to infer probabilities for all snippets.
\acp{stcnn} performance was significantly better than \acp{sfcnn} ($p < 10 ^ {-7}$) (\cref{tab:models}).
The difference between \acp{stcnn} was not significant ($p = 0.012$).

\subsection{Aggregation for seizure-level classification}
\label{sec:exp_agg}

In this experiment, we compared the performance of three aggregation methods to perform seizure-level classification, using 1) the mean, 2) an \ac{rnn} with 64 \ac{lstm} units and 3) an \ac{rnn} with 64 \ac{blstm} units to aggregate the $n$ feature vectors sampled from the video segments.
We used the dataset of feature vectors generated by R(2+1)D-34 (8) (\cref{sec:exp_feat}).
For the task of classifying \ac{fos} and \ac{tcs}, the number of segments should be
selected to ensure snippets after $t_G$, when generalizing semiologies begin, are sampled.
The theoretical minimum number of segments needed to sample snippets after $t_G$ is $n\st{min} = \lceil 1 / (1 - r\st{max}) \rceil $,
where $r\st{max}$ is the largest possible ratio of non-generalizing to generalizing seizure durations (\cref{sec:dataset}).
We can estimate $r\st{max}$ from our dataset:
$r\st{max} = \max ( r_1, \dots, r_{n\st{TCS}} ) = 0.93$, where $n\st{TCS}$ is the number of \acp{tcs},
which yields $n\st{min} = 15$ segments.
We evaluated model performance using $n \in \{2, 4, 8, 16\}$ segments per video and a sampling distribution using $\gamma \in \{ 1, 1.25, 2, 4, \infty \}$, corresponding to uniform, near semi-elliptic, parabolic, near Gaussian and Dirac's delta distributions, respectively.
For evaluation, we used $\gamma \rightarrow \infty$, i.e., only the central snippet of each segment.
We trained using mini-batches with sequences sampled from 64 videos, and an initial learning rate of $10 ^ {-2}$.
We used a weighted binary cross-entropy loss for training,
where the weight for \acp{tcs} was the ratio of \acp{fos} to \acp{tcs}.

\begin{figure}[b!]
    \centering
    \includegraphics[width=\linewidth, trim=0 7 0 7, clip]{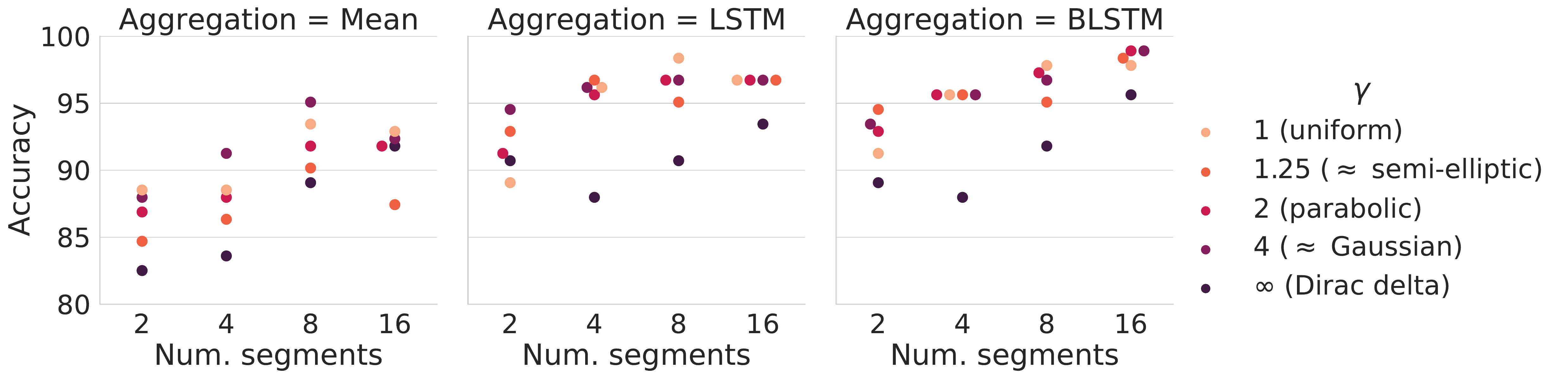}
    \caption{%
        Quantitative results for seizure-level classification.
        Marker brightness is proportional to the dispersion associated with the probability distribution used to sample snippets from the video segments (see \cref{fig:betas}).
    }
    \label{fig:aggregation}
\end{figure}

The highest accuracies were obtained using $n = 16$ segments, $\gamma \in \{ 2, 4 \}$ and the \ac{blstm} aggregator (\cref{fig:aggregation}).
The model with the highest accuracy (98.9\%) and $F_1$-score (98.7\%) yielded 77 true positives, 104 true negatives, 2 false positives and 0 false negatives, where \ac{tcs} is the positive class (\cref{sec:dataset}). See the supplementatry materials for examples of videos classified correctly and incorrectly, with different levels of confidence.

\section{Discussion and conclusion}

Objective assessment of seizure semiology from videos is important to determine appropriate treatment for the diagnosed epilepsy type and help reduce \ac{sudep} risk.
Related works focus on \ac{ez} localization by averaging classifications of short snippets from multiple seizures, ignoring order of semiologies, and are not robust to variations seen in real world datasets such as changes in the \ac{fov}.
Moreover, their performance is limited by the size of the training datasets, which are small due to the expense of curating datasets.
Methods that take into account the sequential nature of semiologies and represent the entirety of seizures are needed.

We presented \ac{gesture}, a method combining \acp{tsn} and \acp{rnn} to model long-range sequences of seizure semiologies. 
\ac{gesture} can classify seizures into \acp{fos} and \acp{tcs} with high accuracy.
To overcome the challenge of training on limited data, we used a network pre-trained on large \ac{har} datasets to extract relevant features from seizure videos, highlighting the importance of transfer learning in medical applications.
\ac{gesture} can take videos from multiple cameras, which makes it robust to patients being out of the \ac{fov}.

\comrem{
We found that \acp{stcnn} were better feature extractors than \acp{sfcnn} (\cref{tab:models}).
Accuracy of snippets classification was proportional to the snippet duration.
As in~\cite{ghadiyaram_large-scale_2019}, we did not observe a large difference in performance between the \acp{stcnn} trained using snippets of 8 and 32 frames.
Therefore, we selected the former model for extracting snippet features, as it is less computationally expensive.
Computational cost becomes especially relevant if data augmentation is used during training or if the model is trained end-to-end, which was not the case in this study.

Although we tried to perform a fair comparison between the \acp{sfcnn} and \acp{stcnn} by matching number of layers and parameters, we recognize the effect the domain of the training set has for transfer learning; \acp{sfcnn} were trained on ImageNet, a large dataset of photographs of which only a small fraction contain humans.
We observed overfitting when training with Wide R2D-50-2, probably due to the larger number of output features with respect to R2D-34, which leads to a classifier with more parameters.
However, it seems clear that models designed to capture human motion are more suitable for characterizing seizure semiology.
}

In \cref{sec:exp_feat} we compared \acp{stcnn} to \acp{sfcnn} for snippet-level classification.
To make comparisons fair, we selected models with a similar number of layers (R2D-34) or parameters (Wide R2D-50-2).
We found the larger \ac{sfcnn} had worse performance, due to overfitting to the training dataset.
Classification accuracy was proportional to snippet duration (\cref{tab:models}), meaning that both \acp{stcnn} outperformed \acp{sfcnn}.
We selected R(2+1)D-34 (8) for the aggregation experiment (\cref{sec:exp_agg}), as performance between the two \acp{stcnn} was similar and this model is less computationally expensive.

Using \ac{lstm} or \ac{blstm} units to aggregate features from snippets improved accuracy compared to averaging (\cref{fig:aggregation}), confirming that modeling the order of semiologies is important for accurate seizure representation.
Model performance was proportional to the number of temporal segments, with more segments providing a denser sampling of seizure semiologies. 
Ensuring some dispersion in the probability distributions used to sample snippets improved classification.
One of the two false positives was caused by the patient being out of the \ac{fov} in one of the video streams.
We did not observe overfitting to unrelated events in the videos, such as nurses in the room, to predict \ac{tcs}, and models correctly discriminated between \acp{tcs} and hyperkinetic \acp{fos}.

We demonstrated that methods designed for \ac{har} can be adapted to learn deep representations of epileptic seizures.
This enables a fast, automated and quantitative assessment of seizures.
\Ac{gesture} takes arbitrarily long videos and is robust to occlusions, changes in \ac{fov} and multiple people in the room.
In the future, we will investigate the potential of \ac{gesture} to classify different types of \acp{tcs} and to localize the \ac{ez}, using datasets from multiple \acp{emu}.

\comrem{
Key points
What others have done?
What's new compared to others
Mechanistic implications
Clinical implications
Prognostication
Risk stratification
High-level of risk, more medication and more monitoring
Hyperkinetic seizures in one patient, with nurses, still correctly classified as fos.
Limitations
Limited clinical value?
Only tested on our data
Further work
GCS seizure classification?
Multiview techniques
}

    \section*{Acknowledgments}

This work is supported by the Engineering and Physical Sciences Research Council (EPSRC) [EP/R512400/1].
This work is additionally supported by the EPSRC-funded UCL Centre for Doctoral Training in Intelligent, Integrated Imaging in Healthcare (i4health) [EP/S021930/1] and the Wellcome / EPSRC Centre for Interventional and Surgical Sciences (WEISS, UCL) [203145Z/16/Z].
The data acquisition was supported by the National Institute of Neurological Disorders and Stroke [U01-NS090407].

This publication represents, in part, independent research commissioned by the Wellcome Innovator Award [218380/Z/19/Z/].
The views expressed in this publication are those of the authors and not necessarily those of the Wellcome Trust.

The weights for the 2D and 3D models were downloaded from TorchVision and \myhref{https://github.com/moabitcoin/ig65m-pytorch}, respectively.

    \bibliographystyle{splncs04}
    \bibliography{Sudep}
\end{document}